\documentclass{article}

\usepackage{arxiv}

\usepackage[utf8]{inputenc} 
\usepackage[T1]{fontenc}    
\usepackage{hyperref}       
\usepackage{url}            
\usepackage{booktabs}       
\usepackage{amsfonts}       
\usepackage{nicefrac}       
\usepackage{microtype}      
\usepackage{lineno}
\usepackage{float}
\usepackage{hyperref}
\usepackage{url}

\usepackage{graphicx}

\usepackage{url}
\usepackage{ntheorem}

\usepackage[parfill]{parskip}
\usepackage{multirow}
\usepackage{booktabs} 
\usepackage{commath}
\usepackage{amssymb}
\usepackage[utf8]{inputenc}
\usepackage[english]{babel}
\usepackage{mathtools}

\DeclarePairedDelimiter\floor{\lfloor}{\rfloor}

\usepackage{algorithm}
\usepackage{algorithmic}

\usepackage{tipa}
\usepackage{amssymb}

\newcommand*{\field}[1]{\mathbb{#1}}%

\newtheorem{thm:def}{Definition}[section]
\newtheorem{thm:thm}{Theorem}[section]
\newtheorem{thm:lemma}{Lemma}[section]
\newtheorem{thm:rmk}{Remark}[section]
\newtheorem{thm:corollary}{Corollary}[section]
\newtheorem{thm:pro}{Proposition}[section]
\usepackage{makecell}

\newenvironment{proof}{\paragraph{Proof:}}{\hfill$\square$}

\usepackage[numbers]{natbib}
\title{SPLASH: Learnable Activation Functions for Improving Accuracy and Adversarial Robustness}

\author{
  Mohammadamin Tavakoli\\
  Department of Computer Science\\
  University of California, Irvine\\
  \texttt{mohamadt@uci.edu} \\
   \And
 Forest Agostinelli \\
  Department of Computer Science\\
  University of California, Irvine\\
  \texttt{fagostin@uci.edu} \\
  \And
  Pierre Baldi\\
  Department of Computer Science\\
  University of California, Irvine\\
  \texttt{pfbaldi@uci.edu}\\
}
\date{}

\begin{document}

\maketitle

\begin{abstract}
We introduce SPLASH units, a class of learnable activation functions shown to simultaneously improve the accuracy of deep neural networks while also improving their robustness to adversarial attacks. SPLASH units have both a simple parameterization and maintain the ability to approximate a wide range of non-linear functions. SPLASH units are: 1) continuous; 2) grounded ($f(0) = 0$); 3) use symmetric hinges; and 4) the locations of the hinges are derived directly from the data (i.e. no learning required). Compared to nine other learned and fixed activation functions, including ReLU and its variants, SPLASH units show superior performance across three datasets (MNIST, CIFAR-10, and CIFAR-100) and four architectures (LeNet5, All-CNN, ResNet-20, and Network-in-Network). Furthermore, we show that SPLASH units significantly increase the robustness of deep neural networks to adversarial attacks. Our experiments on both black-box and open-box adversarial attacks show that commonly-used architectures, namely LeNet5, All-CNN, ResNet-20, and Network-in-Network, can be up to 31\% more robust to adversarial attacks by simply using SPLASH units instead of ReLUs.
\end{abstract}

\section{Introduction}
Nonlinear activation functions are fundamental for deep neural networks (DNNs). They determine the class of functions that DNNs can implement and influence their training dynamics, thereby affecting their final performance. 
For example, DNNs with rectified linear units (ReLUs) \citep{nair2010rectified} have been shown to perform better than logistic and tanh units in several scenarios \citep{pedamonti2018comparison, nwankpa2018activation, nair2010rectified, goodfellow2016deep}. 
Instead of using a fixed activation function, one can use a parameterized activation function and learn its parameters to add flexibility to the model. Piecewise linear functions are a reasonable choice for the parameterization of activation functions \citep{agostinelli2014learning, He_2015, ramachandran2017searching, jin2016deep, li2016multi} due to their straightforward parameterization and their ability to approximate non-linear functions \cite{garvin1957applications, stone1961approximation}. However, in the context of deep neural networks, the best way to parameterize these piecewise linear activation functions is still an open question. Previous piecewise linear activation functions either sacrifice expressive power for simplicity (i.e. having few parameters) or sacrifice simplicity for expressive power. While expressive power allows deep neural networks to approximate complicated functions, simplicity can make optimization easier by adding useful inductive biases and reducing the size of the hypothesis space. Therefore, we set out to find a parameterized piecewise linear activation function that is as simple as possible while maintaining the ability to approximate a wide range of functions.

Piecewise linear functions, in the most general form, are real-valued functions defined as $S+1$ line segments with $S$ hinges that denote where one segment ends and the next segment begins. As detailed in Section \ref{sec:PLtoSPLASH}, a function of this most general form requires $3S+2$ parameters. Many functions in this hypothesis space, such as discontinuous functions, are unlikely to be useful activation functions. Therefore, we significantly reduce the size of the hypothesis space while maintaining the ability to approximate a wide range of useful activation functions. We restrict the form of the piecewise linear function to be continuous and grounded (having an output of zero for an input of zero) with symmetric and fixed hinges. By doing so, we reduce the number of parameters to $S+1$. Furthermore, we still maintain the ability to approximate almost every successful deep neural network activation function. We call this parameterized piecewise linear activation function SPLASH (\textbf{S}imple \textbf{P}iecewise \textbf{L}inear and \textbf{A}daptive with \textbf{S}ymmetric \textbf{H}inges). 


Typically, learned activation functions are evaluated in terms of accuracy on a test set. We compare the classification accuracy of SPLASH units to nine other learned and fixed activation functions and show that SPLASH units consistently give superior performance. We also perform ablation studies to gain insight into why SPLASH units improve performance and show that the flexibility of the SPLASH units during training significantly affects the final performance. In addition, we also evaluate the robustness of SPLASH units to adversarial attacks \cite{szegedy2013intriguing,goodfellow2014explaining,nguyen2015deep}. When compared to ReLUs, SPLASH units reduce the success of adversarial attacks by up to 31\%, without any modifications to how they are parameterized or learned.

\section{Related Work}
Variants of ReLUs, such as leaky-ReLUs \citep{maas2013rectifier}, exponential linear units (ELUs) \citep{clevert2015fast}, and scaled exponential linear units (SELUs) \citep{klambauer2017self} have been shown to improve upon ReLUs. ELUs and SELUs encourage the outputs of the activation functions to have zero mean while SELUs also encourage the outputs of the activation functions to have unit variance. 
Neural architecture search \citep{ramachandran2017searching} has also discovered novel activation functions, in particular, the Swish activation function. The Swish activation function is defined as $f(x) = x*sigmoid(\beta x)$ and performs slightly better than ReLUs. It is worth mentioning that, in \citet{lin2013network}, the authors proposed the network-in-network approach where they replace activation functions in convolutional layers with small multi-layer perceptrons. Theoretically, due to universal approximation theorem \citep{csaji2001approximation}, this is the most expressive activation function; however, it requires many more parameters.

Some of the early attempts to learn activation functions in neural networks can be found in \citet{poli1996parallel}, \citet{weingaertner2002hierarchical}, and \citet{khan2013fast}, where the authors proposed learning the best activation function per neuron among a pool of candidate activation functions using genetic and evolutionary algorithms. Maxout \citep{goodfellow2013maxout} has been introduced as an activation function aimed at enhancing the model averaging properties of dropout \citep{srivastava2014dropout}. However, not only is it limited to approximating convex functions, but it also requires a significant increase in parameters.

APL units \cite{agostinelli2014learning}, P-ReLUs \citep{He_2015} and S-ReLUs \citep{jin2016deep} are adaptive activation functions from the piecewise linear family that can mimic both convex and non-convex functions. Of these activation functions, APL units are the most general. However, they require a parameter for the slope of each line segment as well as for the location of each hinge. Additionally, APL units give more expressive power to the left half of the input space than to the right half. Furthermore, the locations of the hinges are not determined by the data and, therefore, it is possible that some line segments may go unused. S-ReLUs also learn the slopes of the line segments and the locations of the hinges, however, the initial locations of the hinges are determined by the data. S-ReLUs have less expressive power than APL units as the form of the function is restricted to only have two hinges. P-ReLUs are the simplest of these activation functions with one fixed hinge where only the slope of one of the line segments is learned. On the other hand, SPLASH units can have few or many hinges and the the locations of the hinges are fixed and determined by the data. Therefore, only the slopes of the line segments have to be learned. Furthermore, SPLASH units give equal expressive power to the left and the right half of the input space.

\section{From Piecewise Linear Functions to SPLASH Units}
\label{sec:PLtoSPLASH}
\subsection{Family of Piecewise Linear Functions}
Given $S+1$ line segments and $S$ hinges, piecewise linear functions can be parameterized with $2(S+1)+S=3S+2$ parameters: one parameter for the slope and one parameter for the y-intercept of each segment, plus $S$ parameters for the locations of the hinges. We reduce the number of parameters to $S+1$ while still being able to approximate a wide range of functions by restricting the activation function to be \textbf{continuous} and \textbf{grounded} with \textbf{symmetric} and \textbf{fixed} hinges.

\subsubsection*{Continuous}
The general form of piecewise linear functions allows for discontinuous functions. Because virtually all successful activation functions are continuous, we argue that continuous learnable activation functions will still provide sufficient flexibility for DNNs. For a continuous piecewise linear function, we need to specify the y-intercept of one segment, the slopes of the $S+1$ segments, as well as the locations of the $S$ hinges, reducing the number of parameters to $2S + 2$.

\subsubsection*{Grounded}
Furthermore, we restrict the function to be grounded, that is, having an output of zero for an input of zero. We can do this without loss of generality as a function that is not grounded can still be created with the use of a bias. Since the y-intercept is fixed at zero, we no longer have to specify the y-intercept for any of the segments, reducing the number of parameters to $2S+1$.

\subsubsection*{Symmetric Hinges}
In our design, we place the hinges in symmetric locations on the positive and negative halves on the x-axis, giving equal expressive power to each half. This allows, if need be, the activation function to approximate both even and odd functions. Because the location of one hinge determines the location of another, we can reduce the number of parameters for the hinges to $\floor*{\frac{S}{2}}$. In the case of an odd number of hinges, one hinge will be fixed at zero to maintain symmetry. This reduces the number of parameters to $S+1 + \floor*{\frac{S}{2}}$

\subsubsection*{Fixed Hinges}
Finally, we address the issue of where to set the exact location of each segment. It is important that each segment has the potential to influence the output of the function. The distribution of the input could be such that only some of the segments influence the output while others remain unused. In the worst case, the input could be concentrated on a single segment, reducing the activation function to just a linear function. To ensure that each segment is able to play a role in the output of the function, we train our DNNs using batch normalization \cite{ioffe2015batch}. At the beginning of training, batch normalization ensures that, for each batch, the input to the activation function has a mean of zero and a standard deviation of one. Using this knowledge, we can place the hinges at fixed locations that correspond to a certain number standard deviations away from the mean. With the location of the hinges fixed, the number of parameters is reduced to $S+1$. This activation function can approximate the vast majority of existing activation functions, such as tanh units, ReLUs, leaky ReLUs, ELUs,  and, with the use of a bias, logistic units. We show the different types of piecewise linear functions that we have described in Table \ref{tab:plfamily}.

\begin{table}[]
\centering
\begin{tabular}{cc@{\hskip 0.1in}c@{\hskip 0.1in}c@{\hskip 0.1in}c@{\hskip 0.1in}c}
Type          & \makecell{General} & \makecell{Continuous} & \makecell{Continuous \\ Grounded} & \makecell{Continuous \\ Grounded\\Symmetric hinges} & \makecell{Continuous\\Grounded\\Symmetric hinges\\Fixed hinges}\\
\\
\# Params & $3S+2$  & $2S+2$     & $2S+1$                   & $S+1+\floor*{\frac{S}{2}}$ & $S+1$ \\
\\

Viz &  \parbox[c]{7em}{\includegraphics[width=1in]{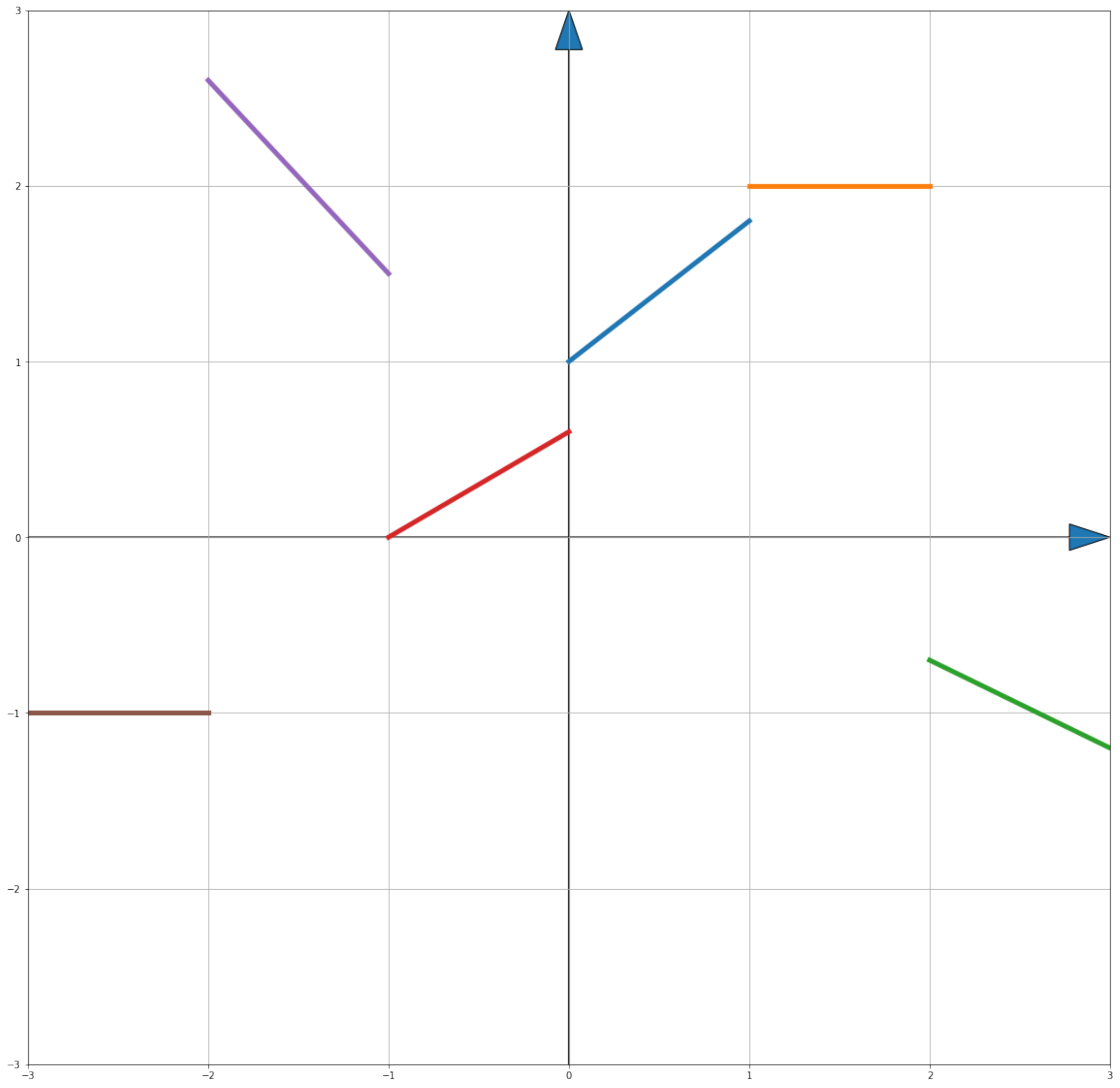}}       &  \parbox[c]{7em}{\includegraphics[width=1in]{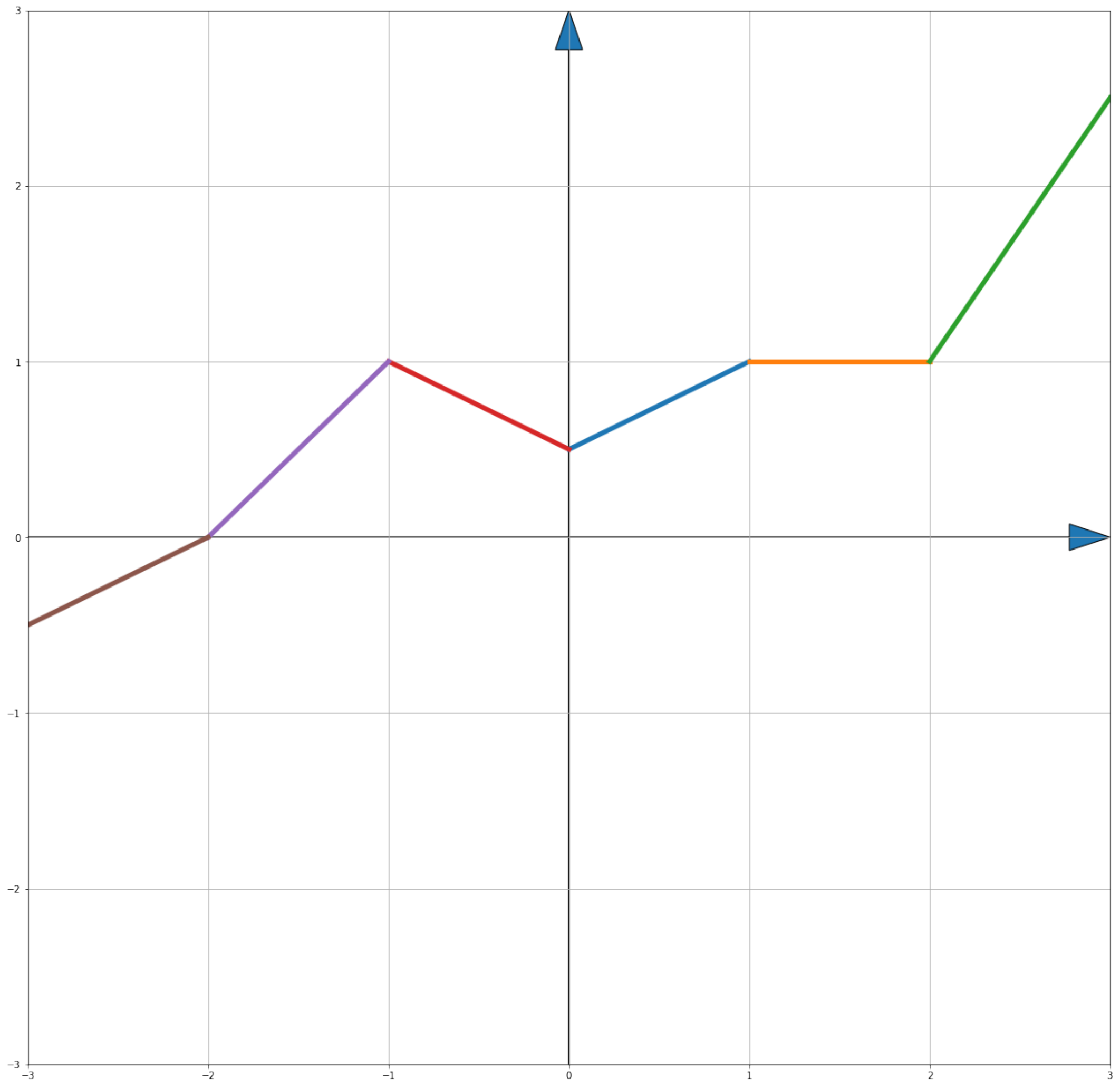}}          &  \parbox[c]{7em}{\includegraphics[width=1in]{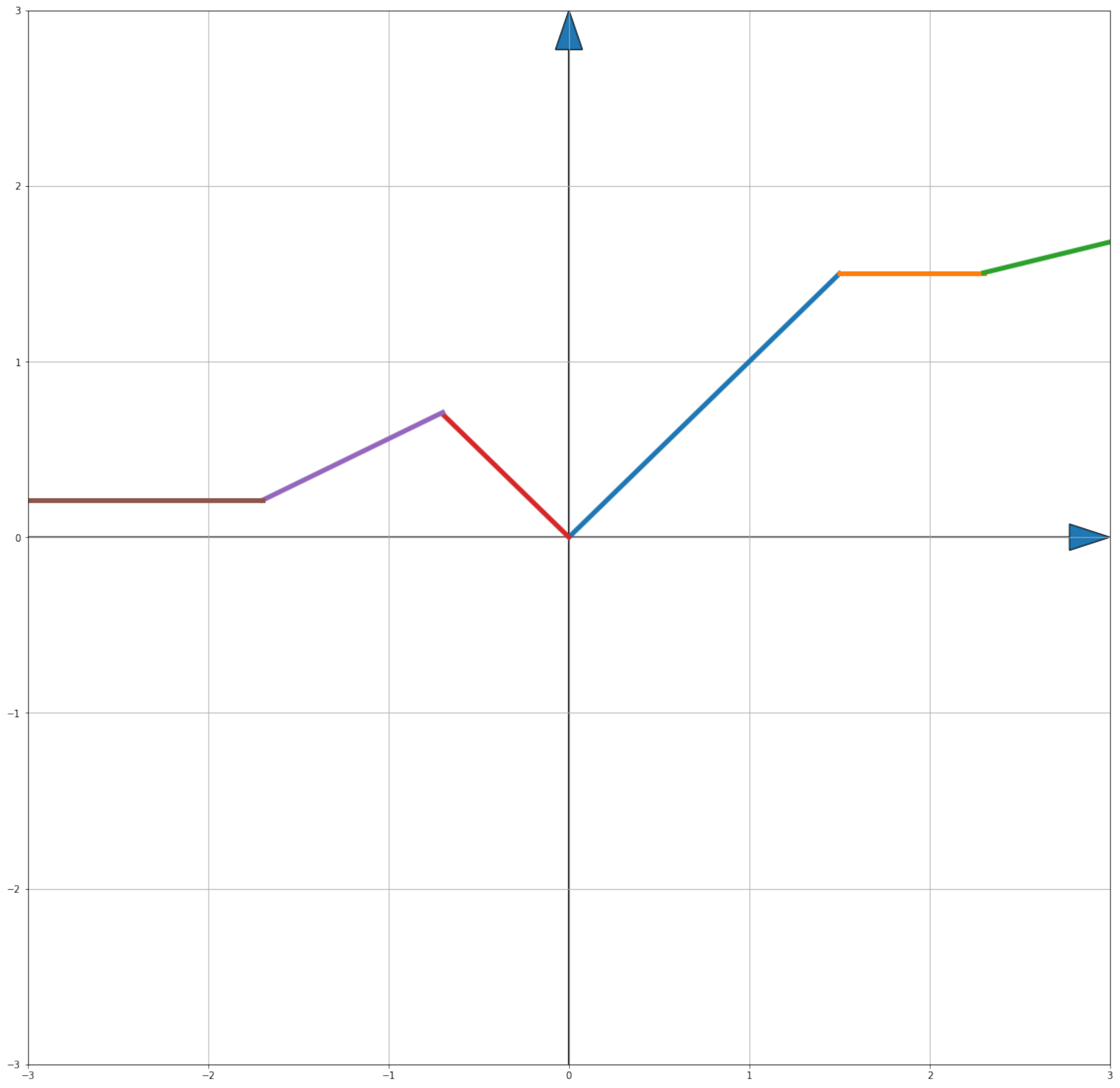}}                         &     \parbox[c]{7em}{\includegraphics[width=1in]{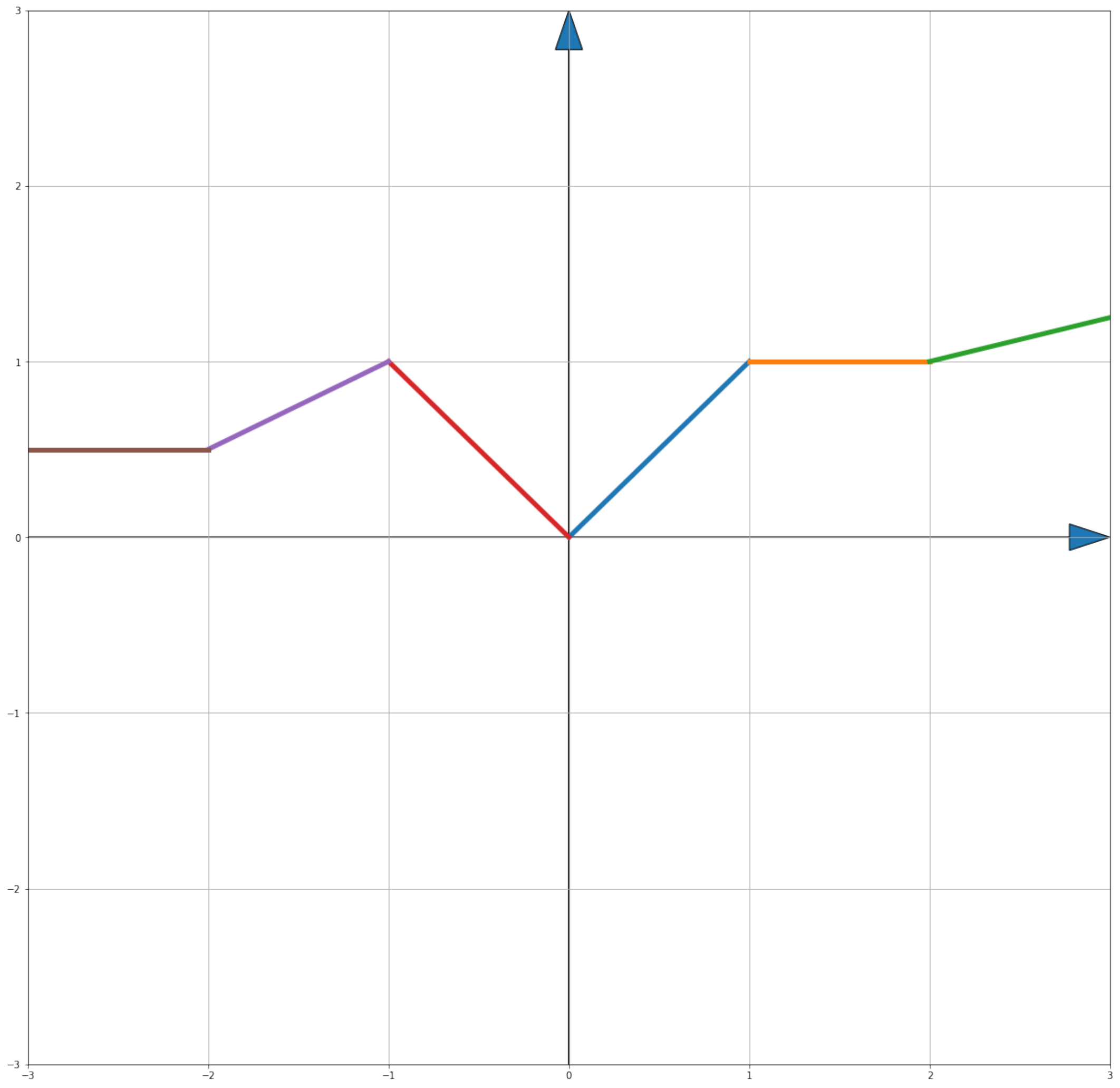}}   &  \parbox[c]{7em}{\includegraphics[width=1in]{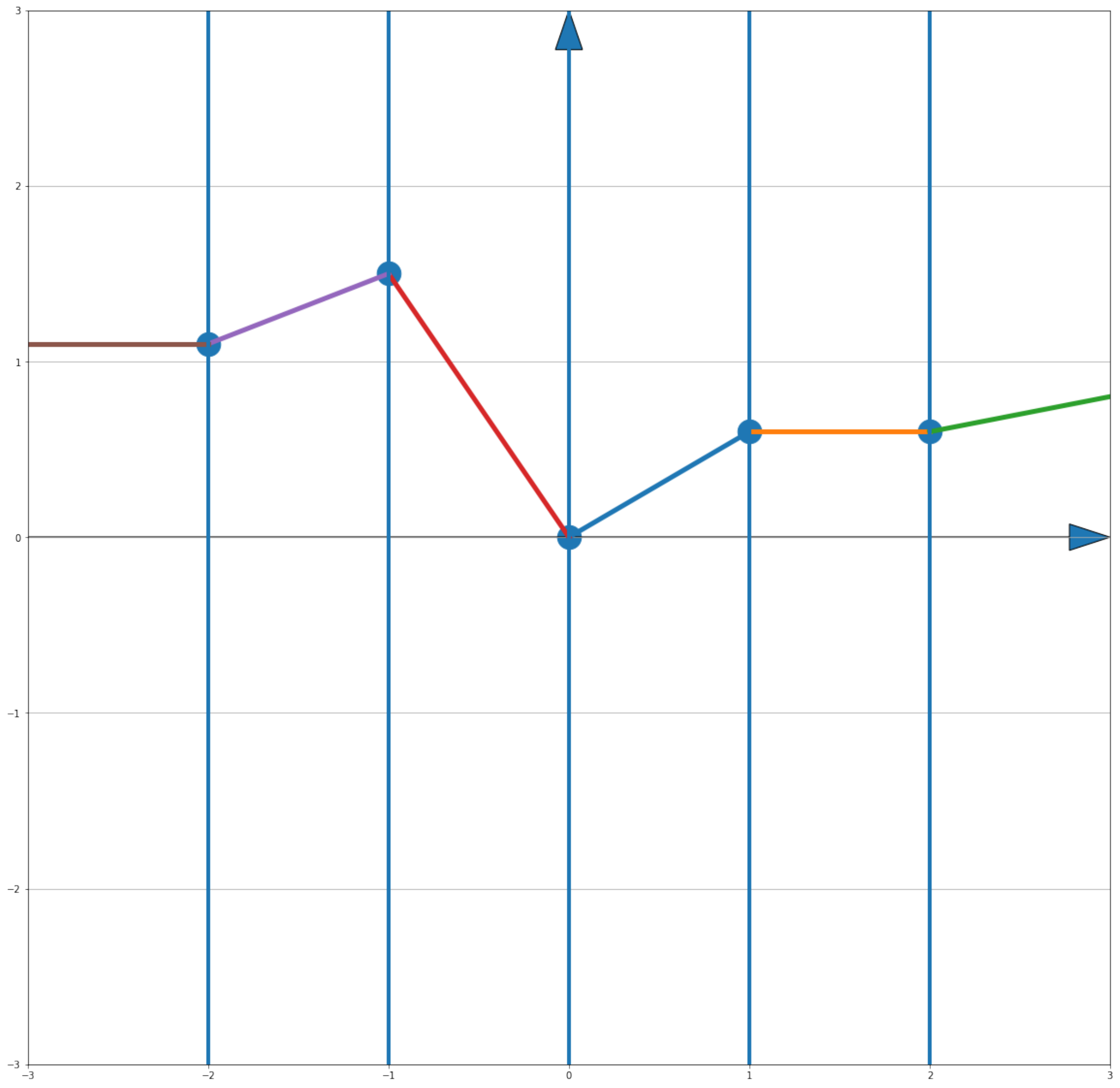}}    

\end{tabular}\\
\caption{Different types of piecewise linear functions defined on $N$ intervals. The rightmost function is what we use to parameterize our SPLASH activation functions.}
\label{tab:plfamily}
\end{table}

\subsection{SPLASH Units}

We formulate the activation of a hidden unit $h$ as the summation of $S+1$ max functions with $S$ symmetric offsets, where $S$ is an odd number and one of the offsets is zero: 
\begin{equation}
    \label{eqn:eqn2}
    h(x) = \sum_{s=1}^{(S+1)/2}a_{+}^smax(0, x - b^s) +\sum_{s=1}^{(S+1)/2}a_{-}^smax(0, -x - b^s)
\end{equation}

The first summation contains max functions with a non-zero output starting at $x\ge0$ and continuing to infinity. The second summation contains max functions with a non-zero output starting at $x\le0$ and continuing to negative infinity. When summed together, these max functions form $S+1$ continuous and grounded line segments with hinges located at $b^s$ and $-b^s$. To ensure the function has symmetric and fixed hinges, we use the same $b^s$ in both summations, where $b^s \ge 0$ for all $s$; furthermore, we have the values of $b^s$ remain fixed during training. Since we are using batch normalization, we fix the positions of the hinges $b^s$ for each $s$ to be a predetermined number of standard deviations away from the mean. We ensure there is always one hinge at zero by setting $b^1$ to be zero. The learned parameters $a_{+}^s$ and $a_{-}^s$ determine the slope of each line segment and are shared across all units in a layer. Therefore, SPLASH units add $S+1$ parameters per layer. We study the effect of different initializations as well as the effect of the number of hinges, $S$, on training accuracy. From our experiments, we found that initializing SPLASH units to have the shape of a ReLU and setting $S$ to $7$ gave the best results. More details are given in the appendix.

The following theorem shows that SPLASH units can approximate any non-linear and uniformly continuous function that has an output of zero for an input of zero in a closed interval of real numbers.
\vspace{2mm}
\begin{thm:thm}
    \label{thm:in-r}
    For any function $f: [A, B] \rightarrow \field{R}$ and $\epsilon \in \field{R}^{+}$,  $\exists S \in \field{N}$, where $|f(x)-\text{SPLASH(x)}| \leq \epsilon$, assuming:
        \begin{itemize}
            \item $A$ and $B$ are finite real numbers.
            \item $f$ is uniformly continuous.
        \end{itemize}
    \begin{proof}
    Uniform continuity of $f$ implies that for every $\epsilon \in \field{R}^{+}$, $\exists \delta > 0$ such that for every $x$ and $y$ $\in [A, B]$ where $|x-y| \leq \delta$, then we have $|f(x)-f(y)| \leq \epsilon$.
    Placing $S$ equally distanced hinges $\{H_1, ..., H_S\}$ on the interval $[A, B]$,  divides this into $S+1$ equal sub-intervals $[H_i, H_{i+1}]$. We choose $S$ to be greater than $\frac{B-A}{\delta}-1$, so the length of each sub-interval would be smaller than $\delta$. For any of the sub-intervals starting at $H_i \in \{H_1, ..., H_S\}$, we approximate $f$ by a line segment which connects $f(H_i)$ to $f(H_i+\frac{B-A}{S+1})$. Due to the linear form of SPLASH(x) for $x \in [H_i, H_i+ \frac{B-A}{S+1}]$:
        \begin{align}
        \label{eq4}
         |f(x)-\text{SPLASH(x)}| \leq max(max_{x}|f(x) - f(H_i)|, max_{x}(|f(x)-f(H_{i+1})|)
        \end{align}
    $f$ is uniformly continuous, so:
        \begin{align}
        \label{eq5}
            |f(x)-\text{SPLASH(x)}| \leq \epsilon
        \end{align}

    Now we need to show that SPLASH function (i.e., Equation \ref{eqn:eqn2}) is able to connect $f(H_i)$ to $f(H_i+\frac{B-A}{S+1})$ for $H_i\in\{H_1, ..., H_S\}$.
    We do so by a simple induction as follows:
    Suppose that $f(H_i)$ connected to $f(H_{i+1})$ for $i\in\{1, ..., i-1\}$.
    The slope of SPLASH in the sub-interval $[H_{i-1}, H_{i}]$ are set to be $\sum_{1}^{i}a^i_+$ or $\sum_{1}^{i}a^i_-$ (depending on the sign of the sub-interval). However, the slope of SPLASH in the sub-interval $[H_{i}, H_{i+1}]$ is either $\sum_{1}^{i+1}a^i_+$ or $\sum_{1}^{i+1}a^i_-$. In both cases, the extra term $a^{i+1}_+$ or $a^{i+1}_-$ can change the slope to any arbitrary value. This fact plus the assumption of continuity of SPLASH guarantees that $f(H_{i}) \text{  can be connected to  } f(H_{i+1})$ which was our proposed approximation. The last thing to mention is that since SPLASH is grounded (SPLASH(0)=0), this approximation by line segments can only approximate functions $f$ where $f(0)=0$. 
    \end{proof}
\end{thm:thm}


\section{Accuracy}
\subsection{Comparison to Other Activation Functions}
In order to show that SPLASH units are beneficial for deep neural networks, we compare it with well-known activation functions in different architectures. We train LeNet5, Network-in-Network, All-CNN, and ResNet-20, on three different datasets: MNIST \cite{lecun1998gradient}, CIFAR-10, and CIFAR-100 \cite{krizhevsky2009learning}. We set $S=7$ and fix the locations of the hinges at $x = -2.5, -2, -1, 0, +1, +2, +2.5$. $a_{+}^1$ is initialized to $1$ and the remaining slopes are initialized to $0$. With this initialization, the starting shape of a SPLASH unit mimics the shape of a ReLU.

With the exception of the All-CNN architecture, moderate data augmentation is performed as it is explained in \citet{he2016deep}. Moderate data augmentation adds horizontally flipped examples of all images to the training set as well as random translations with a maximum translation of 5 pixels in each dimension. For the All-CNN architecture, we use heavy data augmentation which is introduced in \citet{springenberg2014striving}. More details on the hyperparameters are mentioned in the appendix.

We compare SPLASH units to ReLUs, leaky-ReLUs,  PReLUs, APL units, tanh units,  sigmoid units,  ELUs,  maxout units with nine features,  and Swish units. We tune the hyperparameters for each DNN using ReLUs and use the same hyperparameters for each activation function. The results of the experiments are shown in Table \ref{tab:allarchs}. We report the average and the standard deviation of the error rate on the test set across five runs. The table shows that SPLASH units have the best performance across all datasets and architectures.

\begin{table}[!]
    \small
    \centering
    \caption{Deep neural networks with ReLUs, leaky-ReLUs, PReLUs, tanh units, sigmoid units, ELUs, maxout units with nine features, Swish units, APL with $S=5$, and SPLASH units are compared on three different datasets. We compare the error rates of SPLASH units on the test set to the \textbf{best} of the other activation functions. For the sake of brevity, D-A refers to Data Augmentation. The values in the tables are error-rates and are reported in percentages. The numbers are shown in the form of \textit{mean}$\pm$\textit{standard deviation}.}
    \vspace{3mm}
    \begin{tabular}{lccccc}
    \toprule
        \multirow{2}{*}{Activation}&\multicolumn{1}{c}{\bf MNIST}&\multicolumn{2}{c}{\bf CIFAR-10}&\multicolumn{2}{c}{\bf CIFAR-100}\\
        \cmidrule(lr){3-4}
        \cmidrule(lr){5-6}
        &  & - & D-A & - & D-A\\
    \midrule
        LeNet5 + ReLU \citep{bigballon2017cifar10cnn} & &31.22&23.77&-&-\\
        LeNet5 (ours) + ReLU &1.11$\pm0.09$& 30.98$\pm1.14$&23.41$\pm1.31$&&\\
        LeNet5 (ours) + PReLU & 1.13$\pm0.04$&30.71$\pm0.69$&23.33$\pm0.88$&&\\
        LeNet5 (ours) + SPLASH & \bf1.03$\pm0.07$&\bf30.14$\pm0.99$&\bf22.93$\pm1.24$&&\\
        \cmidrule(lr){2-6}
        Net in Net + ReLU \citep{lin2013network}  & &10.41&8.81& 35.68&-\\
        Net in Net (ours) + ReLU  & &9.71$\pm0.69$&8.11$\pm0.81$&36.06$\pm0.84$&32.98$\pm1.10$\\
        Net in Net + APL \citep{agostinelli2014learning}  & &9.59$\pm0.24$&7.51$\pm0.14$&34.40$\pm0.16$&30.83$\pm0.24$\\
        Net in Net (ours) + SPLASH & &\bf9.21$\pm0.55$&\bf7.29$\pm0.93$&\bf33.91$\pm0.97$&\bf30.32$\pm0.66$\\
        \cmidrule(lr){2-6}
        All-CNN + ReLU \citep{springenberg2014striving}  & &9.08&7.25&33.71&-\\
        All-CNN (ours) + ReLU  & &9.24$\pm0.48$&7.42$\pm0.59$&34.11$\pm0.79$&32.43$\pm0.73$\\
        All-CNN (ours) + maxout  & &9.19$\pm0.51$&7.45$\pm0.41$&34.21$\pm0.88$&32.33$\pm0.91$\\
        All-CNN (ours) + SPLASH  & &\bf9.02$\pm0.33$&\bf7.18$\pm0.41$&\bf33.14$\pm0.71$&\bf32.06$\pm0.66$\\
        \cmidrule(lr){2-6}
        ResNet-20 + ReLU \citep{he2016deep} & &-&8.75&-&-\\
        ResNet-20 (ours) + ReLU & &10.65$\pm0.55$&8.71$\pm0.51$&34.54$\pm0.88$&32.63$\pm0.67$\\
        ResNet-20 (ours) + APL &&10.29$\pm0.71$ &8.59$\pm0.58$&34.62$\pm0.79$&32.51$\pm0.81$\\
        ResNet-20 (ours) + SPLASH & &\bf9.98$\pm0.42$&\bf8.18$\pm1.02$&\bf33.97$\pm0.51$&\bf32.12$\pm0.77$\\
        
    \bottomrule
    \end{tabular}
    \label{tab:allarchs}
\end{table}

\subsection{Insights into why SPLASH Units Improve Accuracy}
\begin{figure}[h]
  \includegraphics[width=\linewidth, height=70mm]{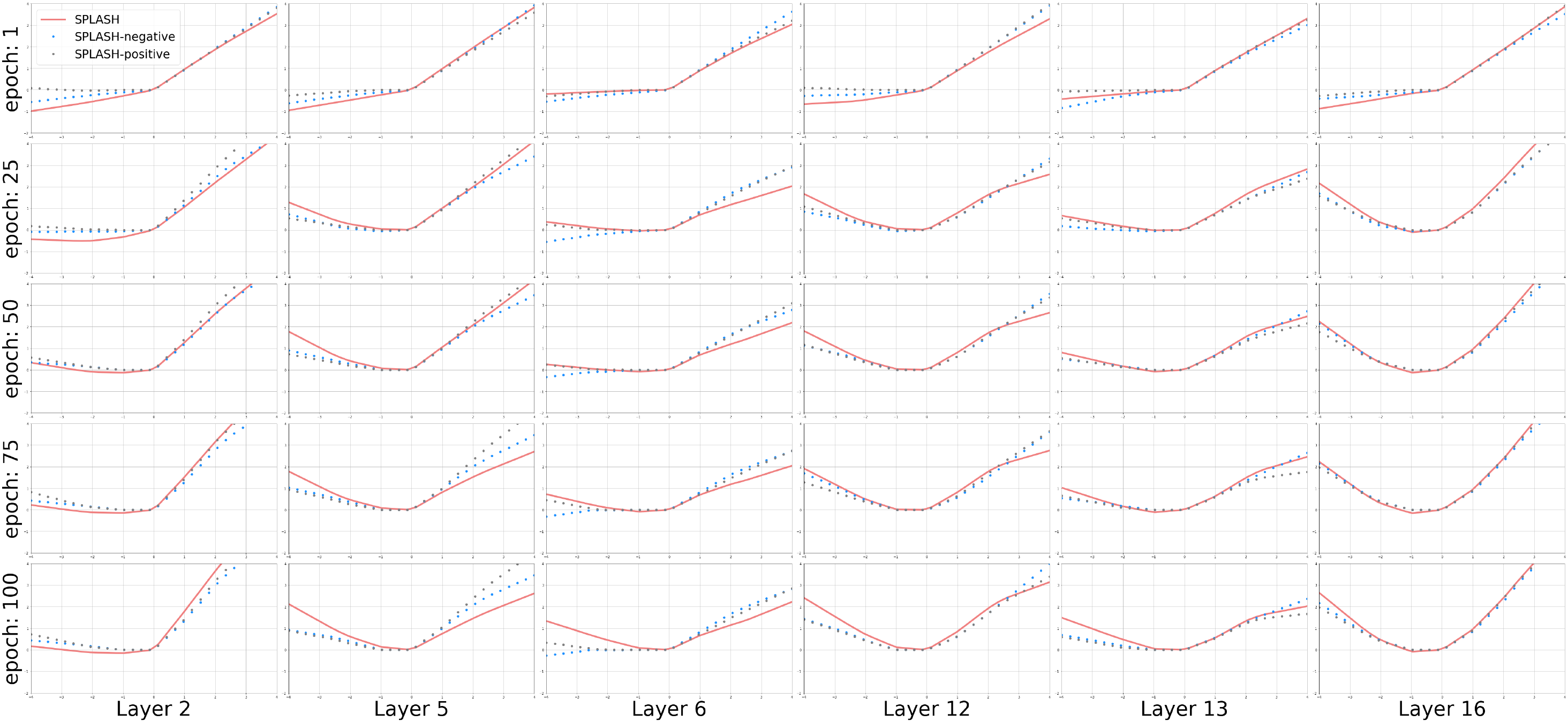}

  \caption{The shape of the SPLASH units in six different layers of the ResNet-20 architecture during training on the CIFAR-10 dataset. In the early stages of training, the shape of SPLASH units looks visually similar to that of a leaky-ReLU. However, during the later stages of training, the shape of SPLASH units looks visually similar to that of a symmetric function.}
  \label{fig:shapeCIFAR}
\end{figure}

Figure \ref{fig:shapeCIFAR} shows how the shape of the SPLASH units change during training for the ResNet-20 architecture. From these figures, we can see that, during the early stages of training, the SPLASH units have a negative output for a negative input and a positive output for a positive input. During the later stages of training, SPLASH units have a positive output for both a negative input and a positive input. SPLASH units look similar to that of a leaky-ReLU during the early stages of training and look similar to a symmetric function during the later stages of training.

\begin{figure}[h]
\small
  \includegraphics[width=\linewidth]{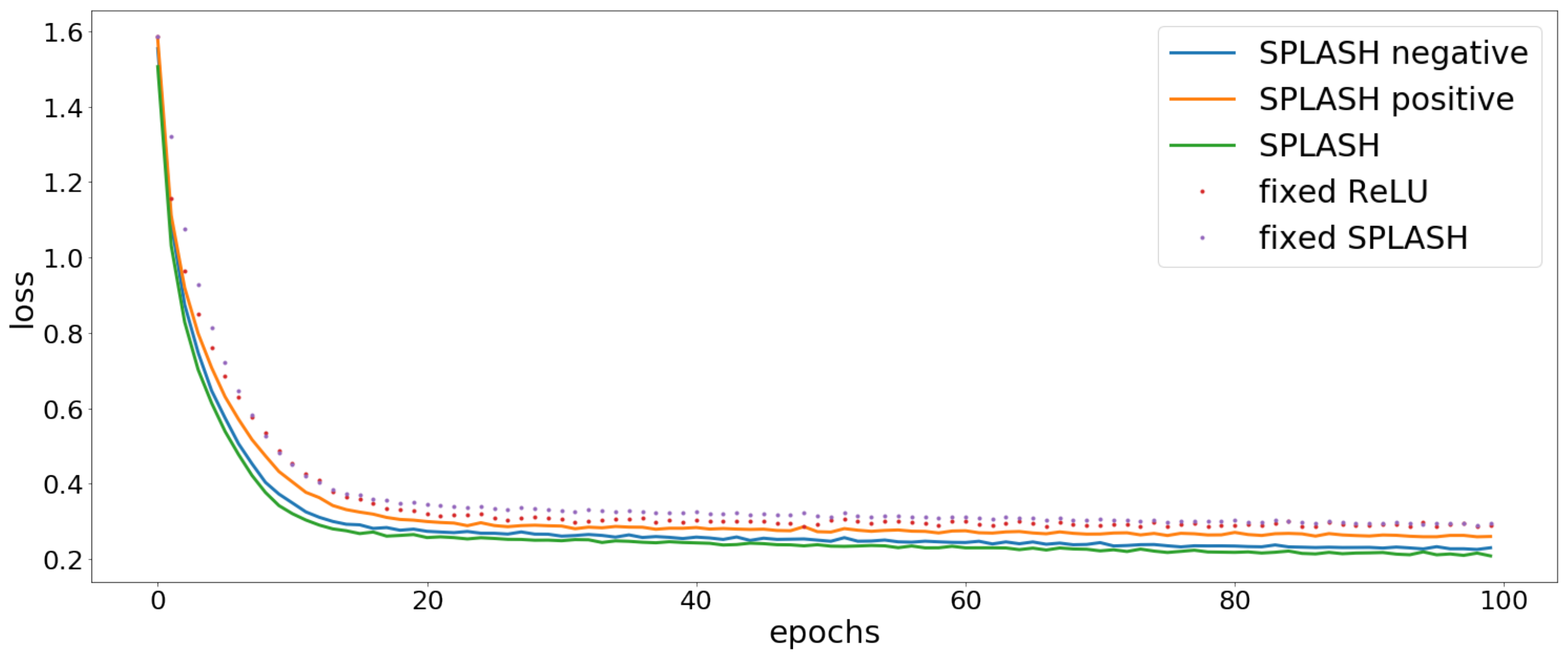}

  \caption{Training loss for ReLUs and different types of SPLASH units for the ResNet-20 architecture on CIFAR-10. SPLASH units converge faster and also have the lowest final loss. Fixed SPLASH is a fixed activation function that mimics the final shape of the SPLASH units trained on the ResNet-20 architecture. Fixed SPLASH performs only about as well as ReLUs. SPLASH-negative units perform better than SPLASH-positive units, however, they perform worse than SPLASH units. Furthermore, although SPLASH-positive units have the ability to mimic the final shape of SPLASH units, they perform worse.}
  \label{fig:loss_inits}
\end{figure}

To better understand why SPLASH units lead to better performance, we used the final shape of the SPLASH units as a fixed activation function to train another ResNet-20 architecture. In Figure \ref{fig:loss_inits}, we can see that the performance is only as good as that of ReLUs. This leads us to believe that the evolution of the shape of the SPLASH units during training is crucial to obtaining improved performance. Since we observed that SPLASH units would first give a negative output for a negative input and then give a positive output for a negative input, we train ResNet-20 with SPLASH units under two different conditions: 1) the first slope on the negative half of the input ($a_{-}^{1}$) is forced to be only positive, yielding a negative output for the line segment at zero (SPLASH-negative units) and 2) the first slope on the negative half of the input ($a_{-}^{1}$) is forced to be only negative, yielding a positive output for the line segment at 0 (SPLASH-positive units).

The performance of SPLASH-positive and SPLASH-negative units is shown in Figure \ref{fig:loss_inits}. The figure shows that, although SPLASH-positive units have the ability to mimic the final learned shape of SPLASH units, it performs worse than SPLASH units and only slightly better than ReLUs. This shows that the ability to give a negative output for a negative input is crucial for SPLASH units. Furthermore, SPLASH-negative units perform better than SPLASH-positive units, but still worse than SPLASH units. In addition, we see that SPLASH-negative units exhibit a relatively quick decrease in the training loss, similar to that of SPLASH units, but do not reach the final training loss of SPLASH units. These observations suggest that the flexibility of the learnable activation function plays a crucial role in the final performance.

\subsection{Tradeoffs}
The benefits of SPLASH units come at the cost of longer training time. The average per epoch training time and the final accuracy of a variety of fixed and learned activation functions are reported in Table \ref{run-time}. The table shows that training with SPLASH units can take between 1.2 and 3 times longer, depending on $S$ and the chosen architecture. We see that accuracy does not significantly decrease beyond $S=7$. Therefore, we chose $S=7$ for our experiments. While, for many deep learning algorithms, obtaining better performance often comes at the cost of longer training times, in Section \ref{sec:robustness}, we show that SPLASH units also improve the robustness of deep neural networks to adversarial attacks.
\begin{table}[!]
 \small
    \centering
    \caption{Per-epoch training time is reported in seconds. The benefits of SPLASH come at the cost of slower training time. All models are trained using NVIDIA TITAN V GPU with 12036MiB memory and 850MHz. Maxout is trained with six features and APL is set to have five hinges. For the sake of brevity, T and E are corresponding to per-epoch training time and error rate respectively.}
    \vspace{3mm}
    \begin{tabular}{cccccccccccc}
    \toprule
        \multirow{2}{*}{\bf Activation}&&\multicolumn{5}{c}{\bf SPLASH}&\multicolumn{1}{c}{\bf Tanh}& \multicolumn{1}{c}{\bf Maxout}&\textbf{ReLU}&\textbf{Swish}&\textbf{APL}\\
        \cmidrule(lr){3-7}
        & &$S=3$&$S=5$&$S=7$&$S=9$&$S=11$&&&&&\\
    \midrule
        
        \multirow{2}{*}{MNIST (MLP)}&T&10&14&16&18&19&8&13&6&7&14\\
        &E&1.57&1.33&1.13&1.10&1.12&1.88&1.45&1.35&1.35&1.40\\
        
     \midrule   
        
        \multirow{2}{*}{CIFAR-10 (LeNet5)}&T&21&24& 29 &33&35&19&22&17&17&24\\
        &E&30.79&30.57&30.20 &30.14&30.11&31.14&31.01&30.88&30.69&30.66\\
    \bottomrule
    \end{tabular}
    \label{run-time}
\end{table}

\section{Robustness to Adversarial Attacks}
\label{sec:robustness}
DNNs have been shown to be vulnerable to many types of adversarial attacks \citep{szegedy2013intriguing,goodfellow2014explaining}. Research suggests that activation functions are a major cause of this vulnerability \citep{zantedeschi2017efficient, brendel2017decision}. For example, \citet{zhang2018efficient} bounded a given activation function using linear
and quadratic functions with adaptive parameters and applied a different activation for each neuron to make neural networks robust to adversarial attacks. 
\citet{wang2018adversarial} proposed a data-dependent activation function and empirically showed its robustness to both black-box and gradient-based adversarial attacks.
Other studies such as \citet{rakin2018defend}, \citet{dhillon2018stochastic}, and \citet{song2018defense} focused on other properties of activation functions, such as quantization and pruning, and showed that they can improve the robustness of DNNs to adversarial examples.

Recently, authors in \citet{zhao2016suppressing} theoretically showed that DNNs with symmetric activations are less likely to get fooled.
The authors proved that \textit{``symmetric units suppress unusual signals of exceptional magnitude which result in robustness to adversarial fooling and higher expressibility.''}
Because SPLASH units are capable of approximating a symmetric function, they may also be capable of increasing the robustness of DNNs to adversarial attacks. In this section, we show that SPLASH units greatly improve the robustness of DNNs to adversarial attacks. This claim is verified through a wide range of experiments with the CIFAR-10 dataset under both black-box and open-box methods, including the one-pixel-attack and the fast gradient sign method.

\begin{figure}[!ht]
  \includegraphics[width=\linewidth]{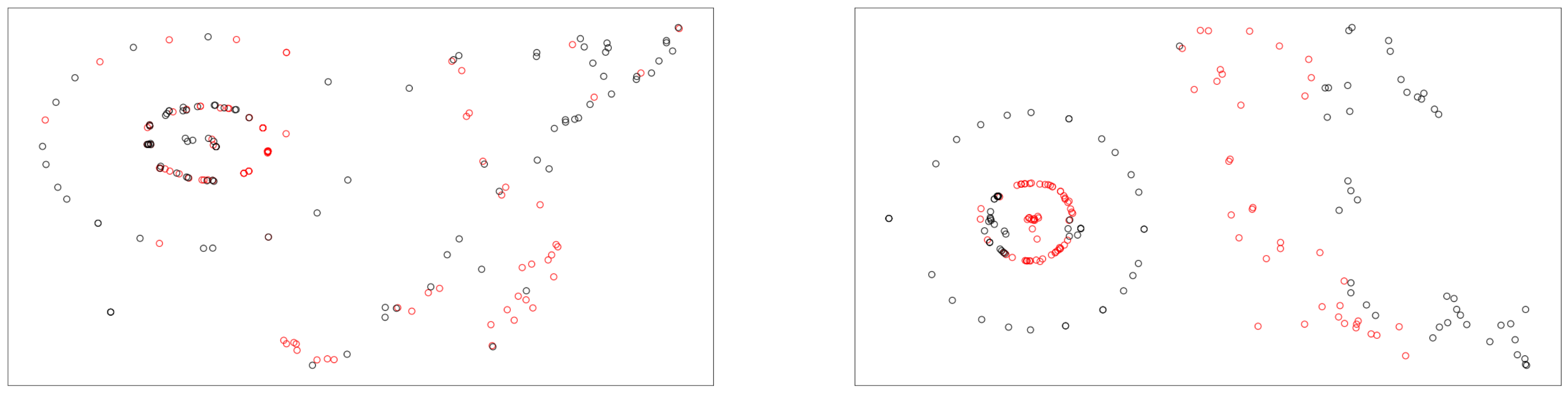}

  \caption{tSNE visualization of the pre-softmax layer's outputs for the LeNet5 architecture trained on CIFAR-10. Left: Trained with ReLUs. Right: Trained with SPLASH units. The figures show that the samples from the frog and ship classes are better separated using the DNN trained with SPLASH units.}
  \label{fig:tsne}
\end{figure}

An intuition for why a DNN with SPLASH units is more robust than a DNN with ReLUs is provided in Figure \ref{fig:tsne}. For each of the two networks, we take 100 random samples of frog and ship images and visualize the pre-softmax representations using the tSNE visualization  \citep{maaten2008visualizing} in Figure \ref{fig:tsne}. The figure shows that the two classes have less overlap for the DNN with SPLASH units than for the DNN with ReLUs. 

\subsection{Black-Box Adversarial Attacks}
For black-box adversarial attacks, we assume the adversary has no information about the parameters of the DNN. The adversary can only observe the inputs to the DNN and outputs of the DNN, similar to that of a cryptographic oracle. We test the robustness of DNNs with SPLASH units using two powerful black box adversarial attacks, namely, the one-pixel attack and the boundary attack.

\subsubsection{One Pixel Attack}
A successful one pixel attack was proposed by \citet{Su_2019}, which is based on differential evolution. Using this technique, we can iteratively generate adversarial images to try to minimize the confidence of the true class. The process starts with randomly modifying a few pixels to generate adversarial examples. At each step, several adversarial images are fed to the DNN and the output of the softmax function is observed. Examples that lowered the confidence of the true class will be kept to generate the next generation of adversaries. New adversarial images are then generated through mutations. By repeating these steps for a few iterations, the adversarial modifications generate more and more misleading images. The last step returns the adversarial modification that reduced the confidence of the true class the most, with the goal being that a class other than the true class has the highest confidence. 

In the following experiment, we modify one, three, and five pixels of images to generate adversarial examples. The mutation scheme we used for this experiment is as follows:
\begin{equation}
\label{eqn2}
x_i^{l+1} = x_{r_1}^{l} + 0.5(x_{r_2}^{l}+ x_{r_3}^{l})
\end{equation}

Where $r_1$, $r_2$, and $r_3$ are three non-equal random indices of the modifications at step $l$. $x_i^{l+1}$ will be an element of a new candidate modification.

To evaluate the effect of SPLASH units on the robustness of DNNs, we employ commonly-used architectures, namely, LeNet5, Network-in-Network, All-CNN, and ResNet-20. Each architecture is trained with ReLUs, APL units, Swish units, and SPLASH units. The results are shown in Table \ref{tab:blackbox}. The results show that SPLASH units significantly improve robustness to adversarial attacks for all architectures and outperform all other activation functions. In particular, for LeNet5 and ResNet-20, SPLASH units improve performance over ReLUs by 31\% and 28\%, respectively. 


\begin{table}[!ht]
    \small
    \centering
    \caption{Robustness to the one-pixel attack using 1000 randomly chosen CIFAR-10 test set images. We attack each architecture five times and report the results in the form of \textit{mean}$\pm$\textit{standard deviation} of the number of successful attacks. 
    The maximum number of iterations for all attacks is set to 40. $avg|Z_{true}-Z_{adv}|$ is computed for the one-pixel attack.}
    \vspace{3mm}
    \begin{tabular}{lccccc}
    \toprule
      Model&Activation&one-pixel&three-pixels&five-pixels&$avg|Z_{true}-Z_{adv}|$\\
    \midrule
        \multirow{4}{*}{LeNet5}&ReLU&736$\pm12.3$&803$\pm12.7$&868$\pm28.9$&0.740\\
        &Swish&701$\pm14.2$&780$\pm17.4$&840$\pm11.0$&0.805\\
        &APL&635$\pm15.9$&709$\pm9.8$&781$\pm17.7$&\bf0.465\\
        &SPLASH&\bf514$\pm17.2$&\bf588$\pm7.4$&\bf651$\pm21.7$&0.540\\
        \midrule
     \multirow{4}{*}{Net in Net}&ReLU&644$\pm16.5$&701$\pm20.0$&769$\pm18.3$&0.621\\
        &Swish&670$\pm28.5$&715$\pm33.8$&760$\pm26.1$&0.419\\
        &APL&521$\pm21.2$&661$\pm19.9$&703$\pm22.6$&0.455\\
        &SPLASH&\bf449$\pm18.6$&\bf530$\pm16.3$&\bf599$\pm23.5$&\bf0.311\\
    \midrule
     \multirow{4}{*}{All-CNN}&ReLU&580$\pm17.2$&661$\pm15.0$&707$\pm25.7$&0.366\\
        &Swish&597$\pm23.5$&630$\pm33.9$&699$\pm34.6$&0.511\\
        &APL&509$\pm25.9$&581$\pm21.2$&627$\pm24.0$&0.295\\
        &SPLASH&\bf471$\pm18.8$&\bf515$\pm25.1$&\bf570$\pm24.2$&\bf0.253\\
    \midrule
         \multirow{4}{*}{ResNet-20}&ReLU&689$\pm28.2$&721$\pm28.2$&781$\pm25.3$&0.551\\
        &Swish&650$\pm17.7$&689$\pm17.0$&730$\pm29.7$&0.601\\
        &APL&579$\pm14.4$&631$\pm15.7$&692$\pm19.4$&\bf0.290\\
        &SPLASH&\bf493$\pm24.3$&\bf544$\pm22.9$&\bf579$\pm21.2$&0.332\\
     \bottomrule
    \end{tabular}
    \label{tab:blackbox}
\end{table}

After observing adversarial samples which are deceiving to DNNs with ReLUs and DNNs with SPLASH units, we found that DNNs with SPLASH units still assign higher confidence to the true labels of the perturbed images than ReLUs and Swish units. More precisely, we measure the average of $|Z(x')_{\text{true\_label}}-Z(x')_{\text{adversarial\_label}}|$ over all adversarial samples where both networks are fooled, where $Z(.)$ is the output of the softmax layer and $x'$ is the adversarial sample. For each model, this measurement is included in Table \ref{tab:blackbox}. The results show that SPLASH units often have a smaller average value, again showing that SPLASH units are more robust to adversarial attacks.

\subsubsection{Boundary Attacks}
We use another black-box adversarial attack to further examine the effect SPLASH units have on the robustness of DNNs to adversarial fooling. Boundary attacks, which were recently introduced by \citet{brendel2017decision}, are a powerful and commonly used black-box adversarial attack.
Considering the original pair of input image and the corresponding target as $(x, l_x)$ , the attack algorithm is initialized from an adversarial pair of $(x_{adv}^{0}, l_{adv}^{0})$, where $x_{adv}^{0} \sim \mathcal{N}(0, 1)$ s.t. $l_{adv}^{0} \neq l_x$. Then, a random walk is performed $K$ times along the boundary between the adversarial region, $S_{l_{adv}} | \forall x' \in S_{l_{adv}}, l_x' \neq l_x$, and the region of the true label such that (1) $x_{adv}$ stays in the adversarial region and (2) the distance towards the original image $d(x, x_{adv}^{k})$ is reduced.
The random walk uses the following three steps: (1) Draw a random sample $\mu$ from an i.i.d. Gaussian as the direction of the next move. (2) Project the sampled direction onto the sphere centered at $x$ with a radius of $||x-x_{avd}^{k-1}||$ and take a step of size $\epsilon = \frac{||\mu^k||_2}{d(x, x_{adv}^{k-1})}$ in this projected direction. This step guarantees that the perturbed image gets closer to the original image at each step. (3) Make a move of size $\delta$ towards the original image, where $\delta = \frac{d(x, x_{adv}^{k-1})-d(x, x_{adv}^{k})}{d(x, x_{adv}^{k-1})}$. Ideally, this algorithm will converge to the adversarial sample $x_{adv}^{K}$ which is the closest to the original input $x$. The details and hyper-parameters of the attack are explained in the appendix.

In what follows, we employ the same architectures and activation functions that were used in the previous section. The results of this attack are shown in Table \ref{boundary}. We observe that DNNs with SPLASH units are more robust to this adversarial attack than DNNs with APL units, ReLUs, and Swish units. 

\begin{table}[!ht]
    \small
    \centering
    \caption{Robustness to the boundary attack using 1000 randomly chosen CIFAR-10 test set images. We attack each architecture five times and report the results in the form of \textit{mean}$\pm$\textit{standard deviation} of the number of successful attacks.}
    \vspace{3mm}
    \begin{tabular}{lccc}
    \toprule
      Model&Activation& \# of successful attacks&$avg|Z_{true}-Z_{adv}|$\\
    \midrule
        \multirow{4}{*}{LeNet5}&ReLU&801$\pm14.4$&0.815\\
        &Swish&779$\pm9.2$&0.511\\
        &APL&730$\pm12.0$&0.541\\
        &SPLASH&\bf619$\pm15.8$&\bf0.401\\
    \midrule
        \multirow{4}{*}{Net in Net}&ReLU&766$\pm9.0$&0.502\\
        &Swish&759$\pm5.4$&0.391\\
        &APL&654$\pm11.7$&\bf0.340\\
        &SPLASH&\bf598$\pm10.1$&0.351\\
    \midrule
     \multirow{4}{*}{All-CNN}&ReLU&744$\pm9.0$&0.621\\
        &Swish&700$\pm16.2$&0.710\\
        &APL&672$\pm6.5$&0.480\\
        &SPLASH&\bf611$\pm11.9$&\bf0.421\\
    \midrule
         \multirow{4}{*}{ResNet-20}&ReLU&790$\pm6.4$&0.548\\
        &Swish&793$\pm11.3$&0.566\\
        &APL&711$\pm9.2$&0.471\\
        &SPLASH&\bf621$\pm9.4$&\bf0.349\\
     \bottomrule
    \end{tabular}
    \label{boundary}
\end{table}

\subsection{Open-Box Adversarial Attacks}
For open-box adversarial attacks, the adversary now has information about the parameters of the DNN. To further explore the robustness of DNNs with SPLASH units, in this section, we consider two of the popular benchmarks of open-box adversarial attacks: the fast gradient sign method (FGSM) \citep{goodfellow2014explaining} and Carlini and Wagner (CW) attacks \citep{carlini2017towards}. For both attack methods, we consider four different architectures and compare the rate of successful attacks for each of the networks with ReLUs, Swish units, APL units, and SPLASH units. The dataset and architectures are the same as those used for black-box adversarial attacks.

\subsubsection{FGSM}
FGSM generates an adversarial image $x'$ from the original image $x$ by maximizing the loss $L(x', y)$, where $y$ is the true label of the image $x$. This maximization problem is subjected to $||x-x'||_{\infty}\leq\epsilon$ where $\epsilon$ is considered as the \textit{attack strength}. Using the first order Taylor series approximation, we then have:
\begin{equation}
\label{taylo}
L(x', y) = L(x, y) + \nabla_{x} L(x, y)^T . (x-x')
\end{equation}
So the adversarial image $x'$ would be:
\begin{equation}
\label{FGSM}
x' = x + \epsilon.sign(\nabla_{x} L(x, \theta))
\end{equation}

The results for different $\epsilon$ are summarized in Table \ref{tab:dif-ep}. The results show that SPLASH units are consistently better than all other activation functions with performance improvements of up to 28.5\%.

\begin{table}[!ht]
    \small
    \centering
    \caption{Robustness to the FGSM attack using 1000 randomly chosen CIFAR-10 test set images. We attack each architecture five times with random start and report the results in the form of \textit{mean}$\pm$\textit{standard deviation} of the number of successful attacks. $avg|Z_{true}-Z_{adv}|$ is computed for $\epsilon = 0.04$.}
    \vspace{3mm}
    \begin{tabular}{lccccc}
    \toprule
      Model&Activation&$\epsilon=0.02$&$\epsilon=0.04$&$\epsilon=0.06$&$avg|Z_{true}-Z_{adv}|$\\
    \midrule
        \multirow{4}{*}{LeNet5}&ReLU&690$\pm13.5$&755$\pm16.6$&825$\pm24.1$&0.710\\
        &Swish&634$\pm11.1$&740$\pm15.7$&830$\pm25.7$&0.713\\
        &APL&611$\pm22.5$&691$\pm13.4$&807$\pm19.0$&\bf0.419\\
        &SPLASH&\bf493$\pm15.1$&\bf598$\pm21.4$&\bf772$\pm26.7$&0.521\\
    \midrule
         \multirow{4}{*}{Net in Net}&ReLU&590$\pm12.9$&651$\pm17.5$&798$\pm17.1$&0.609\\
        &Swish&577$\pm12.1$&619$\pm14.6$&750$\pm15.3$&\bf0.439\\
        &APL&531$\pm20.4$&607$\pm19.6$&719$\pm20.3$&0.561\\
        &SPLASH&\bf498$\pm12.6$&\bf554$\pm17.4$&\bf689$\pm11.4$&0.499\\
    \midrule

          \multirow{4}{*}{All-CNN}&ReLU&561$\pm10.5$&653$\pm18.1$&741$\pm24.4$&0.590\\
        &Swish&519$\pm12.1$&622$\pm17.3$&740$\pm16.6$&0.576\\
        &APL&522$\pm18.4$&615$\pm8.5$&721$\pm21.7$&0.549\\
        &SPLASH&\bf479$\pm11.2$&\bf588$\pm14.1$&\bf676$\pm19.6$&\bf0.333\\
    \midrule
         \multirow{4}{*}{ResNet-20}&ReLU&651$\pm18.1$&736$\pm16.1$&801$\pm20.7$&0.641\\
        &Swish&639$\pm18.4$&730$\pm17.3$&793$\pm19.7$&0.522\\
        &APL&609$\pm9.4$&701$\pm18.0$&749$\pm20.5$&\bf0.303\\
        &SPLASH&\bf541$\pm16.4$&\bf617$\pm21.7$&\bf711$\pm21.0$&0.411\\

     \bottomrule
    \end{tabular}
    \label{tab:dif-ep}
\end{table}

\begin{table}[!ht]
    \small
    \centering
    \caption{Robustness to the CW-L2 attack using 1000 randomly chosen CIFAR-10 test set images. We attack each architecture five times and report the results in the form of \textit{mean}$\pm$\textit{standard deviation} of the number of successful attacks.}
    \vspace{3mm}
    \begin{tabular}{lccc}
    \toprule
      Model&Activation&\# of successful attacks &$avg|Z_{true}-Z_{adv}|$\\
    \midrule
        \multirow{4}{*}{LeNet5}&ReLU&932$\pm5.5$&0.801\\
        &Swish&919$\pm6.4$&0.713\\
        &APL&922$\pm7.5$&0.609\\
        &SPLASH&\bf898$\pm6.4$&\bf0.541\\
    \midrule
    
         \multirow{4}{*}{Net in Net}&ReLU&916$\pm8.0$&0.790\\
        &Swish&919$\pm5.4$&0.724\\
        &APL&915$\pm6.1$&\bf0.653\\
        &SPLASH&\bf892$\pm5.5$&0.674\\
     \midrule
     \multirow{4}{*}{All-CNN}&ReLU&894$\pm13.7$&0.611\\
        &Swish&887$\pm8.6$&0.631\\
        &APL&876$\pm12.1$&0.509\\
        &SPLASH&\bf863$\pm11.7$&\bf0.365\\
    \midrule
         \multirow{4}{*}{ResNet-20}&ReLU&903$\pm11.8$&0.603\\
        &Swish&911$\pm15.1$&\bf0.441\\
        &APL&894$\pm11.5$&0.590\\
        &SPLASH&\bf870$\pm12.3$&0.541\\

     \bottomrule
    \end{tabular}
    \label{tab:cwl2}
\end{table}

\subsubsection{CW-L2}

Another open-box adversarial attack, which is generally more powerful than FGSM, was introduced in \citet{carlini2017towards}. For a given image $x$ and label $y$, this technique tries to find the minimum perturbation $\delta$, so that the perturbed image $x'$ is classified as $t \neq y$. Using the $L_2$ norm, this perturbation minimization problem can be formulated as follows:
\begin{equation}
\label{cwl2-1}
\forall t \neq y, min ||\delta||^{2}_{2} \hspace{3mm} \text{subject to} \hspace{3mm} f(x+\delta) = t, \hspace{2mm} x+\delta \in [0, 1]^n
\end{equation}

To ease the satisfaction of equality, Equation \ref{cwl2-1} can be rephrased as $min ||\delta||^{2}_{2}+ c.g(x+\delta)$ where $g(x) = max(max_{t \neq y}(logit(x)_{t} - logit(x)_{y}))$, $c$ is Lagrange multiplier, and $logit(x)$ is the pre-softmax vector for the input $x$.

The robustness performance of ReLUs, Swish units, APL units, and SPLASH units for the CW-L2 attack is shown in Table \ref{tab:cwl2}. The table is consistent with previous results as it shows that SPLASH units are the most robust to this adversarial attack.

\section{Conclusion}
SPLASH units are simple and flexible parameterized piecewise linear functions that simultaneously improve both the accuracy and adversarial robustness of DNNs. They had the best classification accuracy across three different datasets and four different architectures when compared to nine other learned and fixed activation functions. When investigating the reason behind their success, we found that the final shape of the learnable SPLASH units did not serve as a good non-learnable (fixed) activation function. Additionally, in our ablation studies, we saw that restricting the flexibility of the activation function hurts performance, even if the restricted activation function can still mimic the final shape of the unrestricted SPLASH units. It could be possible that changes in the activation functions play a particular role in shaping the loss landscape of deep neural networks \cite{hochreiter1997flat, dauphin2014identifying, choromanska2015loss}. Future work will use visualization techniques \cite{craven1992visualizing, gallagher2003visualization, li2018visualizing} to obtain an intuitive understanding of how learnable activation functions affect the optimization process.

Though no adversarial examples are shown during training, SPLASH units still significantly increase the robustness of DNNs to adversarial attacks. Prior research suggests that the reason for this may be related to their final shape, which looks visually similar to that of a symmetric function \citep{zhao2016suppressing}. Given that research has shown that certain activation functions may make deep neural networks susceptible to adversarial attacks \cite{croce2018randomized}, it is possible that adding more inductive biases aimed at reducing these vulnerabilities may increase the robustness of learned activation functions to adversarial attacks. Since our ablation studies have shown the importance of having flexible activation functions during training, these inductive biases may need to allow for flexibility or be applied during the later stages of training, for example, in the form of a regularization penalty.

\section{Acknowledgement}
Work in part supported by ARO grant 76649-CS, NSF grant 1839429, and NSF grant NRT 1633631 to PB. We wish to acknowledge Yuzo Kanomata for computing support.

\bibliography{main}

\section{Appendix}

\subsection{Initialization of SPLASH weights}
\begin{figure}[!ht]
  \includegraphics[width=\linewidth, height=40mm]{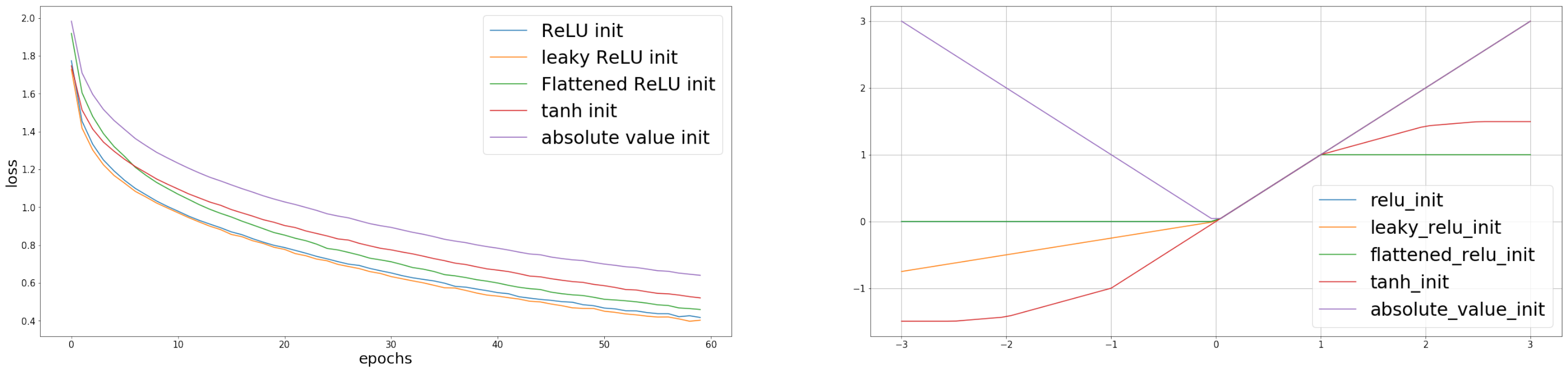}

  \caption{Left: The loss trajectory of training LeNet5 architecture on CIFAR-10 using different initializations of SPLASH units. Right: Visualizations of the initializations.}
  \label{fig:diff_inits}
\end{figure}

In order to choose the best initialization of SPLASH weights ($a_i$ and $b_i$), we compare the performance of five different LeNet5 architecture trained on CIFAR-10. Each of these architectures uses a differently initialized SPLASH activation. Figure \ref{fig:diff_inits} shows that the leaky ReLU and ReLU initializations perform the best. Leaky ReLUs require us to determine the slope of the line segment on the left side of the x-axis. Adding another parameter that may possibly need tuning. Therefore, for simplicity, we use the ReLU initialization ($a_+^1=0$, ans all other parameters set to $0$) in all of our experiments.

\subsection{Number of Hinges}

In this section, we perform a variety of experiments to find the best setting for SPLASH activation in terms of both complexity and performance.

\begin{table}[!ht]
\small
    \centering
    \caption{Three classifications tasks are performed with five different numbers of hinges for SPLASH activation. The number of additional parameters due to the use of SPLASH and training loss are compared below. The MLP architecture consists of three layers each with 256, 64, and 32 units. LeNet5 is used for CIFAR-10. For each experiment, two numbers are reported which are corresponding to shared SPLASH units and independent SPLASH units respectively.}
    \vspace{3mm}
    \begin{tabular}{lcccccc}
    \toprule
        S &3&5&7&9&11\\
    \midrule
    \textbf{Error rate} &&&&&\\
    \midrule
    
        MNIST &  1.57-1.61&1.33-1.39& 1.13-1.17& 1.10-1.08&1.12-1.08\\
        CIFAR-10 &30.79-30.55&30.57-30.29&30.20-30.18&30.14-30.22&30.11-30.19\\
    \midrule
    \textbf{\# of additional params} &&&&&\\
    \midrule
        MNIST  &12-1408&18-2112&24-2816&30-3520&36-4224\\
        CIFAR-10 &16- $>$75k&24-$>$120k&32-$>$150k&40-$>$180k &48-$>$225k\\
    \bottomrule
    \end{tabular}
    \label{tab:num_hinges}
\end{table}
First, we assess the effect of $S$ on the performance of SPLASH. Due to Theorem \ref{thm:in-r}, greater $S$ values increase the expressive power of the SPLASH which generally results better training performance.
We tried $S \in \left[3, 5, 7, 9, 11 \right]$, with symmetrically fixed hinges for SPLASH units. We also use MNIST \citep{lecun-mnisthandwrittendigit-2010} and CIFAR-10 \citep{cifar10}. Each network is trained with two types of SPLASH activations; 1) A \textbf{shared SPLASH}: a shared unit among all neurons of a layer and 2) An independent SPLASH unit for each neuron of a layer.
As it is summarized in Table \ref{tab:num_hinges}, in all cases of $S\geq7$ there is no significant improvement in the performance of the DNNs.

On the other hand, due to the increase in the number of parameters of SPLASH, the activation units become more computationally expensive. In Table \ref{run-time} we compare the \textbf{per-epoch training run-time} for different number of hinges of a \textbf{shared SPLASH}.

For small values of $S$, we can see that SPLASH is comparable to an exponential activation unit such as Tanh, and much faster than heavier activation such as Maxout. 

As one can conclude from both Table \ref{tab:num_hinges} and \ref{run-time}, there is a trade-off between the complexity of SPLASH units and the performance of DNNs. We believe that $S=7$ is the best choice for the number of hinges.


\subsection{Experiments' Details and Statistical Significance}
In this section, we explain the experimental conditions and all the parameters used for each experiment. Also, in order to make the results of Table \ref{tab:allarchs} more interpretable, we perform a t-test \citep{kim2015t} on all the error rates achieved in that experiment.

\textbf{In section 4},  experiments corresponding to Table \ref{tab:allarchs} are performed using four different architectures.
LeNet5 is used as it was introduced in \citet{lecun1998gradient}. It has two convolution layers followed by two MLPs that are connected to a softmax layer. We use our own implementation of LeNet5 with all the hyper-parameters from \citet{bigballon2017cifar10cnn}. However, We train the networks for 100 epochs.


All-CNN architecture which is only taking advantage of convolutional layers, was introduced in \citet{springenberg2014striving}. Since we could not reproduce the exact numbers for the top-1 accuracy on CIFAR-10 dataset using the specifications in the main article, we used our own implementation. We use a learning rate of 0.1, with the decay rate of 1e-6 and momentum of 0.9. The batch size is set to 64 and we trained the networks for 300 epochs. The rest of the hyper parameters are the same as those mentioned in \citet{springenberg2014striving}.

For ResNet architectures, we try a popular variant, ResNet-20, introduced in \citet{he2016deep} which has 0.27M parameters.
Our implementation of ResNet-20 is taken from \citet{chollet2015keras} and \citet{bigballon2017cifar10cnn}. All the hyper-parameters including batch size, number epochs, initialization, learning rate and it's decay, and optimizer are left to the default values of the mentioned repositories.

Lastly, Net in Net architecture which is using an MLP instead of a fixed nonlinear transformation is taken from \citet{bigballon2017cifar10cnn}. We use the same set of hyper-parameters including batch size, number of epochs, learning rate, and etc as it was mentioned in \citet{bigballon2017cifar10cnn}.

In Table \ref{tab:pval}, we show the statistical significance of the experiments performed in section 4. Since each number is the average of five experiments, we are able to perform a t-test and provide p-values and statistical significance for each individual experiment. As one can see in Table \ref{tab:pval}, most of the numbers of Table \ref{tab:allarchs} are statistically significant.

\begin{table}[!ht]
    \small
    \centering
    \caption{The best activation among ReLU, leaky-ReLU, PReLU, tanh, sigmoid, ELU, maxout (nine features), Swish: $x.sigmoid(\beta x)$ is chosen by the minimum average of the error rate.  Then the significance of the comparison between the best network and the network with SPLASH activation is calculated through a t-test. The p-vales for each comparison is provided below.}
    \vspace{3mm}
    \begin{tabular}{lccccc}
    \toprule
        \multirow{2}{*}{Activation}&\multicolumn{1}{c}{\bf MNIST}&\multicolumn{2}{c}{\bf CIFAR-10}&\multicolumn{2}{c}{\bf CIFAR-100}\\
        \cmidrule(lr){3-4}
        \cmidrule(lr){5-6}
        & - & - & D-A & - & D-A\\
    \midrule
        LeNet5 (PReLU vs SPLASH) & 0.057&0.043&0.055&&\\
        Net in Net(ReLU vs SPLASH)& &0.042&0.039&0.038&0.055\\
        All-CNN (maxout vs SPLASH) & &0.041&0.050&0.066&0.061\\
        ResNet-20 (PReLU vs SPLASH)&&0.033 &0.044&0.046&0.044\\
        
    \bottomrule
    \end{tabular}
    \label{tab:pval}
\end{table}

\begin{figure}[!ht]
  \includegraphics[width=\linewidth]{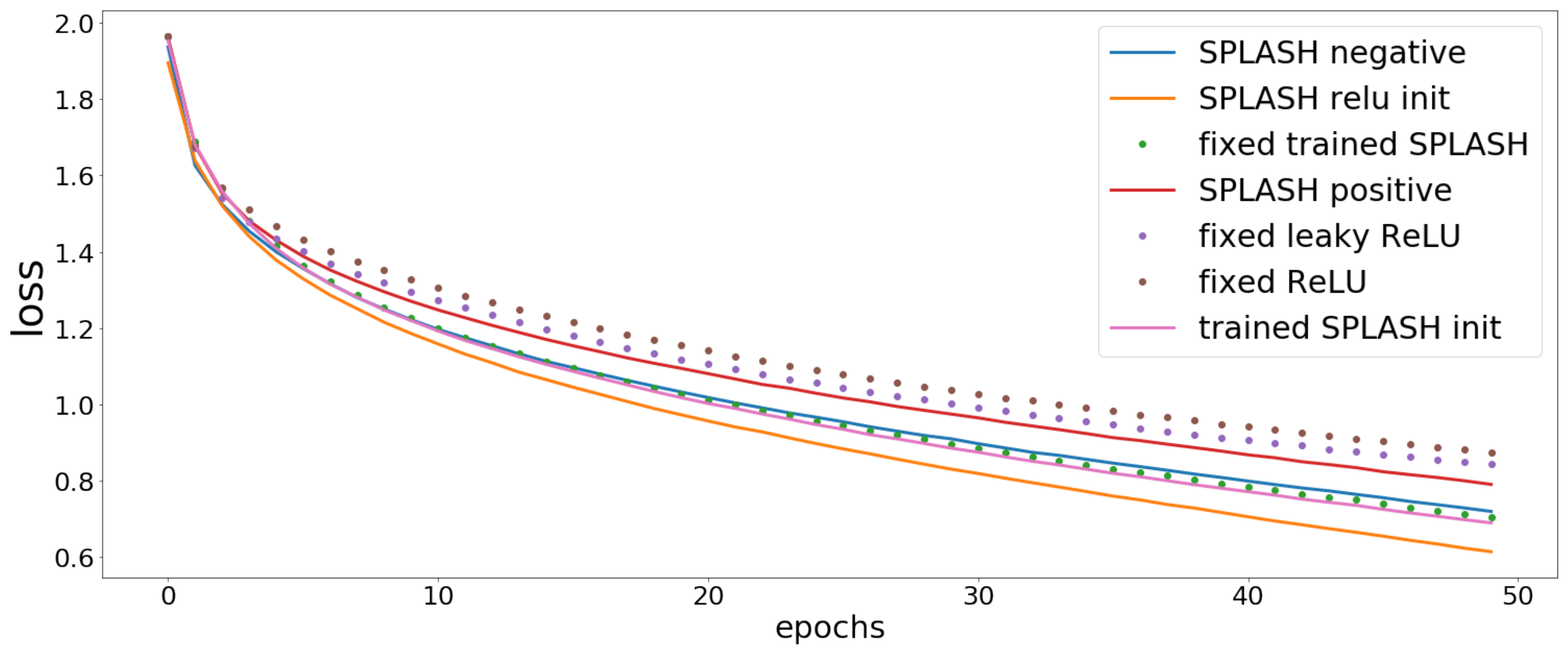}

  \caption{Training loss trajectory for different SPLASH initializations compared to fixed ReLU and leaky ReLU.}
  \label{fig:loss_initsLenet}
\end{figure}

\begin{figure}[!ht]
        \includegraphics[width=\linewidth]{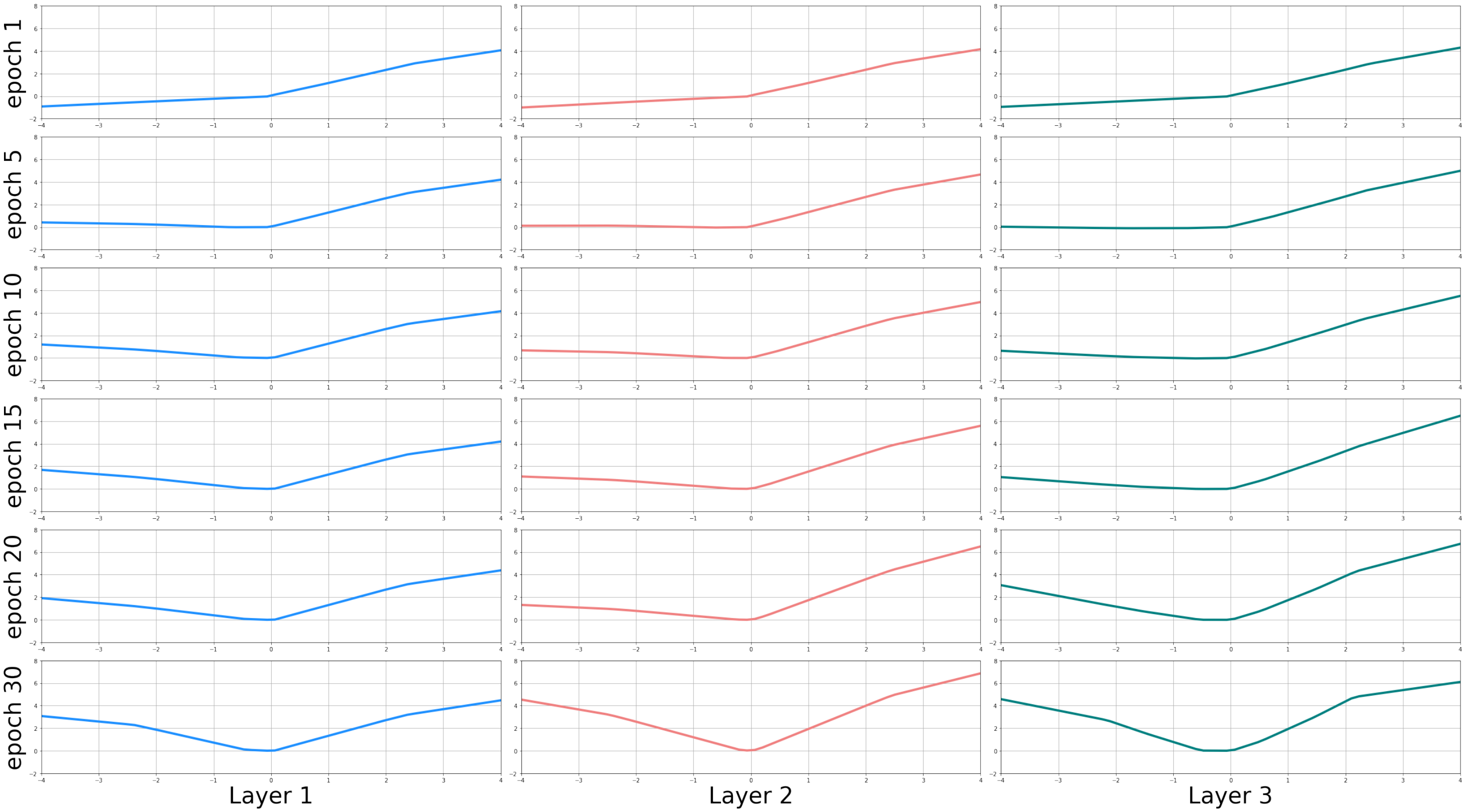}

        \caption{Shape of SPLASH activation during training a simple network of MLPs on MNIST dataset.}
        \label{fig:mlpacitvation}
  
    \end{figure}

\begin{figure}[!th]
  \includegraphics[width=\linewidth, height=80mm]{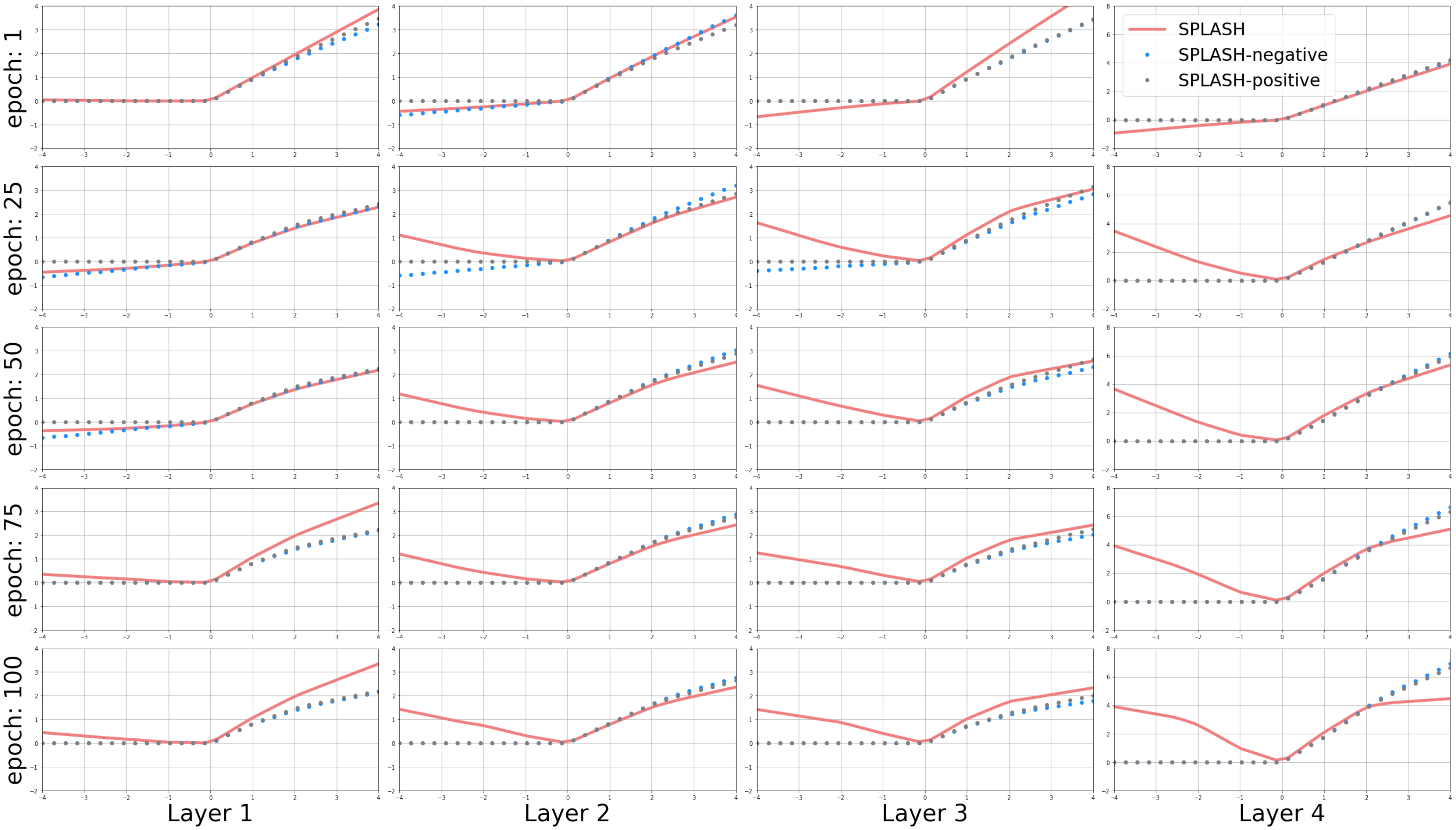}

  \caption{Shape of SPLASH during training a LeNet5 architecture on CIFAR-10 dataset.}
  \label{fig:lenetactivation}
\end{figure}

\textbf{In section 5}, we use ResNet-20 architecture to visualize SPLASH shapes at different stages of the training process. Here we include two more plots showing the evolution of SPLASH units during training. Figure \ref{fig:mlpacitvation} and Figure \ref{fig:lenetactivation} are sowing the evolution of SPLASH units during training MLP and LeNet5 architectures respectively. Both architectures are described in section 4. 

\textbf{In section 6}, we start by a tSNE visualization of 100 random samples of frogs and ships images from the CIFAR-10 test set. The tSNE mapping is performed using a learning rate of 30 and a perplexity of 40.

For the black-box adversarial attack experiments, each network is attacked five times and the reported number is the average of successful modifications in five different attacks. One-pixel-attacks are done using the maximum number of iteration to be 40 and the pop size to be 400. 
For the boundary attack, we use the implementation in \citet{rauber2017foolbox}. To reduce the rate of successful attacks, the hyper-parameters steps is set to 6000. All other hyper-parameters are left as the default from the mentioned implementation. 

As for the open-box attacks, for both FGSM and CW-L2 attack, we employ the implementation and default hyper-parameters in \citet{rauber2017foolbox}. However, to reduce the attack success rate for CW technique, we use 7 and 1000 for variables binary search steps and steps respectively. The network architectures used for experiments in this section, are identical to the architectures used in section 4.

Lastly, four common;y used activation functions were used to train different DNNs in section 6. ReLU ($y=x \text{  for  } x>0, 0 \text{  otherwise  }$), APL ($S=5$, with fixed hinges on $0, \pm 1, \text{and } \pm 2)$, Swish ($y=x.sigmoid(\beta x)$ with $\beta = 0.2)$, and SPLASH (with the configurations mentioned in the previous section) are used.

\end{document}